\documentclass[runningheads]{llncs}
\usepackage{graphicx}
\usepackage{amsmath}
\usepackage{booktabs}

\begin{document}
\title{Adaptive Forgetting Curves for Spaced Repetition Language Learning}
%
%
\author{Ahmed Zaidi \and
Andrew Caines \and
Russell Moore
\and
Paula Buttery \and
Andrew Rice}
\authorrunning{A. Zaidi et al.}
%
\institute{Department of Computer Science and Technology\\
University of Cambridge\\
15 JJ Thomson Ave\\
Cambridge, CB3 0FD\\
United Kingdom\\
\email{\{ahz22,apc38,rjm49,pjb48,acr31\}@cl.cam.ac.uk}\\}
\maketitle              
\begin{abstract}
The \emph{forgetting curve} has been extensively explored by psychologists, educationalists and cognitive scientists alike. In the context of Intelligent Tutoring Systems, modelling the forgetting curve for each user and knowledge component (e.g.\ vocabulary word) should enable us to develop optimal revision strategies that counteract memory decay and ensure long-term retention. In this study we explore a variety of forgetting curve models incorporating psychological and linguistic features, and we use these models to predict the probability of word recall by learners of English as a second language. We evaluate the impact of the models and their features using data from an online vocabulary teaching platform and find that word complexity is a highly informative feature which may be successfully learned by a neural network model. 

\end{abstract}
\section{Introduction}
Optimal human learning techniques have been extensively studied by researchers in psychology \cite{dunlosky2013improving} and computer science \cite{settles2016trainable,zaidi2017curriculum,moore2019skills,zaidiaccurate}. The impact of learning techniques can be measured by how they affect the long-term retention of the learning materials. Measuring retention requires a model of the human forgetting curve, which plots the probability of recall over time. The first version of the forgetting curve was defined by Ebbinghaus \cite{ebbinghaus1885ueber} but has since been developed further by many researchers who have incorporated additional psychologically grounded variations to the model \cite{tabibian2019enhancing,reddy2017accelerating,mozer2019artificial,DBLP:journals/corr/abs-1905-06873,rubin1996one}. The ideal forgetting curve should adapt to learning materials as well as user meta-features (including current ability). In this study we examine the task of vocabulary learning. We investigate a range of linguistically motivated features, meta-features, and a variety of models in order to predict the probability a given learner will correctly recall a particular word.

\section{Method}

We use the Duolingo spaced repetition dataset \cite{DVN/N8XJME_2017} in order to train and evaluate our features and variety of models. The dataset is filtered for English language learners which results in approximately 4.28 million learner-word datapoints. Our models are a modification of the half-life regression model proposed by Settles \& Meeder \cite{settles2016trainable}.


\subsection{Half-Life Regression (HLR)}
The half-life regression model is defined as follows:
\begin{equation}
p = 2^{-\Delta/h}
\end{equation}
where $p$ is the probability of recall, $\Delta$ is the time since last seen (days) and $h$ is the {\it{half-life}} or strength of the learner's memory. We denote the estimated half-life by $\hat{h}_{\Theta}$, and it is defined as:
\begin{equation}
\hat{h}_{\Theta} = 2^{\Theta \cdot \mathbf{x}}
\end{equation}
where $\Theta$ is a vector of weights for the features $\mathbf{x}$. The features of the model are made up of lexeme tags, one tag for each word in the vocabulary (e.g.\ the lexeme tag for word {\it{camera}} is {\it{camera.N.SG}}). The aim of these features is to capture the inherent difficulty of the word.\\

The HLR model is trained using the following loss function:
\begin{equation}
\ell(\mathbf{x};\Theta) = (p - \hat{p}_{\Theta})^2 + (h - \hat{h}_{\Theta})^2 + \lambda||\Theta||^{2}_{2}
\end{equation}
In practice, it was found that optimising for both $p$ and $h$ in the loss function improved the model. The true value of $h$ is defined as $h = \frac{-\Delta}{log(p)}$.

\subsection{HLR with Linguistic/Psychological Features (HLR+)}\label{sec-lingfeatures}

We now expand on the HLR model by adding additional linguistic, psychological and meta-features to $\mathbf{x}$. We refer to this model as HLR+. The features include {\it{word complexity}} scores estimated by a pre-trained model \cite{gooding2019complex}, {\it{mean concreteness}} scores and {\it{percent known}} based on human judgements \cite{brysbaert2014concreteness}, {\it{SUBTLEX}} word frequencies \cite{van2014subtlex} and {\it{user ids}}. 

The motivation for including complexity as a feature is based on the intuition that the more complex the word, the harder it is to remember. Concreteness is included based on previous work showing that concrete words are easier to remember than abstract words because they activate perceptual memory codes in addition to verbal codes \cite{paivio2013imagery}. SUBTLEX is the relative frequency of an English word based on a corpus of 201.3 million words: we hypothesise that more frequent words are more likely to be encountered and reinforced during the time since last seen $\Delta$. Similarly, we expect that `percent known' (the proportion of respondents familiar with each word based on survey data) will correlate with probability of recall. Lastly, we include user id to capture latent behavioural aspects about the learners. 


\subsection{Complexity-based Half-Life Regression (C-HLR+)}
In addition to adding new features, we now describe a new model that modifies the $p$ such that it directly incorporates word complexity. Gooding et al. \cite{gooding2019complex} derived {\it{word complexity}} to express perceived difficulty. We hypothesise that this will correlate with probability of recall. As the complexity of the word rises, the forgetting curve will become steeper. Therefore, the new model is as follows:

\begin{equation}
p = 2^{-\Delta \cdot C_{i}}/h
\end{equation}
where $C$ is the mean complexity for word $i$. We define estimated half-life $\hat{h}_{\Theta}$ as $2^{\Theta \cdot \mathbf{x}}$ where $\mathbf{x}$ is a vector composed of all of the features described in Section \ref{sec-lingfeatures}.

\subsection{Neural Half-Life Regression (N-HLR+)}
Motivated by the recent success of neural networks, we now describe the N-HLR+ model which replaces $\hat{h}_{\Theta} = 2^{\Theta \cdot \mathbf{x}}$ with a neural network. The network can be described as follows:
\begin{equation} 
\hat{h_{\Theta}} = ReLU(\mathbf{x} \cdot \mathbf{w_{1}})\cdot \mathbf{w_{2}}
\end{equation} 
where the network contains a single hidden layer. $\mathbf{x}$ is a vector of input features, $\mathbf{w_{1}}$ is the weight matrix between the inputs and the hidden layer and $\mathbf{w_{2}}$ is the weight matrix between the hidden layer and the output. We use the same loss function as HLR which optimises for both $p$ and $h$. 

\subsection{Evaluation and Implementation}
We use mean absolute error (MAE) of probability of recall for a lexical item as our evaluation metric, in line with previous work \cite{settles2016trainable}. MAE is defined as: $\frac{1}{D}\sum_{D}^{i=1}\left | p - \hat{p_{\Theta}} \right |_{i}$, where $D$ is the total data instances, $p$ and $\hat{p_{\Theta}}$ are the true probability and model estimated probability of recall, respectively.

We divided the Duolingo English data into 90\% training and 10\% test. We trained all non-neural models (e.g.\ HLR, HLR+, C-HLR) using the following parameters which were tuned on the first 500k data points --- learning rate: $0.001$, alpha $\alpha$: $0.01$, $\lambda$: $0.1$. For all neural models (e.g. N-HLR), we used --- learning rate: $0.001$, epochs: $200$, hidden dim: $4$.  

\section{Results and Discussion}
We can see in Table \ref{tbl:results} that HLR+ did not perform much better than HLR. By modifying the loss function to include complexity as a parameter in the C-HLR+ model, we considerably improved the performance of our model. This was in line with our hypothesis that more complex words are forgotten faster and thus are an important feature in modelling the forgetting curve. 

\begin{figure}
\centering
\includegraphics[width=4cm]{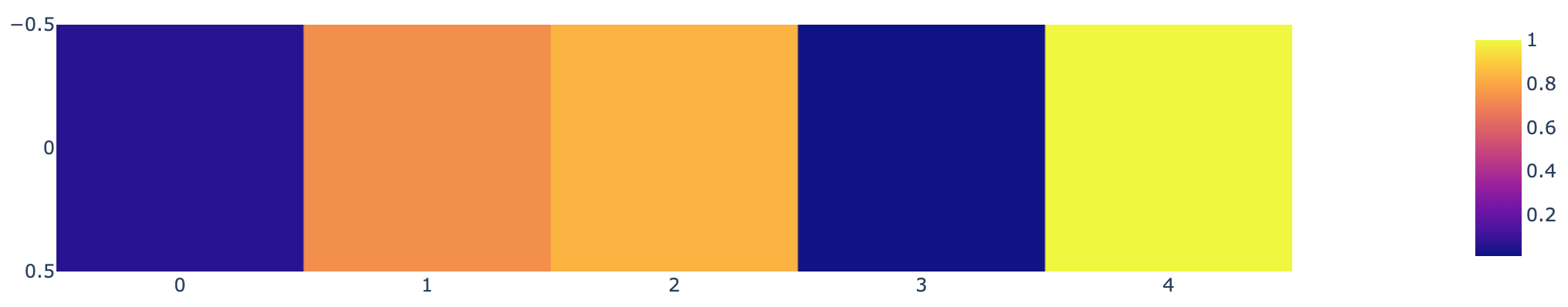}
\caption{A heatmap showing average weights of the hidden layer for the N-HLR+ transformed between $0$ and $1$. Features are in the following order: {\it{user id, concreteness, percent known, SUBTLEX, complexity}}.}   
\label{fig1}
\end{figure}

The N-HLR+ model provided additional improvements to the C-HLR+ model. This is due to the fact that neural models are better at capturing non-linearities between the features and the expected output. Furthermore, when compared to the N-HLR+ model we can see that including complexity into the loss function (CN-HLR+) provides no clear improvements in performance. This is because the model learns to place more importance on the {\it{complexity feature}}. We confirm this by analysing the average weights in the hidden layer of the model as seen in Fig {\ref{fig1}}. The model learns to give greater importance to word complexity, percent known, and concreteness respectively. It does not however, learn much from the user id and SUBTLEX. This is probably due to the fact that a single dimension for capturing user behaviour is not sufficient and that SUBTLEX does not adequately represent learners' experience with English as a second language.

\begin{table}[t]
\caption{\label{tbl:results}Evaluation of forgetting curve models. Pimsleur and Leitner are previous methods of modelling the forgetting curve.}
\begin{center}
\begin{tabular}{lc}
\toprule Model & MAE$\downarrow$ \\
\midrule\midrule
Pimsleur\cite{pimsleur1967memory} & 0.396  \\
Leitner\cite{leitner1972so} & 0.214 \\
Linear Regression & 0.196 \\
HLR\cite{settles2016trainable} & 0.195 \\
HLR-lex\cite{settles2016trainable} & 0.130 \\
\hline
\end{tabular}
\hspace{1em}
\begin{tabular}{lc}
\toprule Model & MAE$\downarrow$ \\
\midrule\midrule
HLR+ & 0.129 \\
C-HLR+ & 0.109  \\
N-HLR+ & \bf 0.105 \\
CN-HLR+ & \bf 0.105 \\
\hline
\end{tabular}
\end{center}
\end{table}

\section{Conclusion}
We present a new model for adaptively learning a forgetting curve for language learning using a modified HLR loss function and a neural network. We incorporate linguistically and psychologically motivated features and show that word complexity is an important feature in predicting probability of recall for a vocabulary item. Furthermore, we illustrate that neural networks can capture the importance of word complexity while a simple HLR fails to take advantage of that signal. This work lays the foundation for work in neural approaches to understanding language learning over time. Future work in this area includes incorporating high-dimensional user embeddings to capture user specific signals that might influence the forgetting curve.

%
%
%
\newpage
\bibliographystyle{splncs04}
\bibliography{main}
%

\end{document}